\documentclass[10pt, a4paper]{article}
\usepackage{lrec2022} 
\usepackage{multibib}
\newcites{languageresource}{Language Resources}
\usepackage{graphicx}
\usepackage{tabularx}
\usepackage{soul}
\usepackage{times}

\usepackage{latexsym, amsmath, amssymb}
\usepackage{amsfonts}
\usepackage{graphicx}
\usepackage{float} 
\usepackage{subfigure} 
\usepackage{url}
\usepackage{algorithm} 
\usepackage{algpseudocode} 
\usepackage{lipsum}
\usepackage{multicol}
\usepackage{titlesec}
\titleformat{\section}{\normalfont\large\bfseries\center}{\thesection.}{1em}{}
\titleformat{\subsection}{\normalfont\SmallTitleFont\bfseries\raggedright}{\thesubsection.}{1em}{}
\titleformat{\subsubsection}{\normalfont\normalsize\bfseries\raggedright}{\thesubsubsection.}{1em}{}
\renewcommand\thesection{\arabic{section}}
\renewcommand\thesubsection{\thesection.\arabic{subsection}}
\renewcommand\thesubsubsection{\thesubsection.\arabic{subsubsection}}

\usepackage{epstopdf}
\usepackage[utf8]{inputenc}

\usepackage{hyperref}
\usepackage{xstring}

\usepackage{color}

\title{ \vspace*{.5\baselineskip} \textbf{Recurrent Neural Networks with Mixed Hierarchical Structures and EM Algorithm for Natural Language Processing}}

\name{Zhaoxin Luo, Michael Zhu} 

\address{Purdue University \\
         \{luo293, yuzhu\}@purdue.edu\\}

\abstract{
   How to obtain hierarchical representations with an increasing level of abstraction becomes one of the key issues of learning with deep neural networks. A variety of RNN models have recently been proposed to incorporate both explicit and implicit hierarchical information in modeling languages in the literature. In this paper, we propose a novel approach called the latent indicator layer to identify and learn implicit hierarchical information (e.g., phrases), and further develop an EM algorithm to handle the latent indicator layer in training. The latent indicator layer further simplifies a text's hierarchical structure, which allows us to seamlessly integrate different levels of attention mechanisms into the structure. We called the resulting architecture as the EM-HRNN model. Furthermore, we develop two bootstrap strategies to effectively and efficiently train the EM-HRNN model on long text documents. Simulation studies and real data applications demonstrate that the EM-HRNN model with bootstrap training outperforms other RNN-based models in document classification tasks. The performance of the EM-HRNN model is comparable to a Transformer-based method called Bert-base, though the former is much smaller model and does not require pre-training. 
 \\ \newline  }

\begin{document}

\maketitleabstract

\section{Introduction}
Text classification is the process of assigning tags or categories to texts according to their contents and is one of the major tasks in Natural Language Processing (NLP) with broad applications such as sentiment analysis, topic labeling, and spam detection.
An important intermediate step in text classification is text representation learning. Previous work uses various neural network models to learn text representation, including Convolution Neural Networks (CNNs) \cite{zhang2015character}, Recurrent Neural Networks (RNNs) \cite{schmidhuber1991neural}, and attention mechanisms \cite{yang2016hierarchical}.

Recently, how to obtain hierarchical representations with an increasing level of abstraction becomes one of the key issues of learning in deep neural networks. 
A variety of hierarchical RNNs have been proposed to incorporate hierarchical representations in modeling languages in the literature. \cite{yang2016hierarchical} proposed to incorporate existing explicit text hierarchical information. In particular,  they proposed to process documents at two levels, which are the word- and sentence-levels, respectively, and obtained promising results.

Another approach to modeling hierarchical and temporal representations is to use multiscale RNNs \cite{schmidhuber1992learning,el1996hierarchical,koutnik2014clockwork}. \cite{chung2016hierarchical} proposed the Hierarchical Multiscale Recurrent Neural Networks (HM-RNNs) equipped with boundary detectors that can discover underlying hierarchical structures without prior information.

\cite{luo2021recurrent} called hierarchical structures with and without prior information as static and dynamic hierarchical structures (or boundaries), respectively, and proposed to use RNNs with Mixed Hierarchical Structure (MHS-RNN) to accommodate both types of structures. In particular, MHS-RNN was used to model documents with word-, phrase-, and sentence-layers, among which the word- and sentence-layers are static and the phrase-layer is dynamic. Further more, 
\cite{luo2021recurrent} added attention mechanism to MHS-RNN to improve its performance. The MHS-RNN model with attention mechanism provides efficient representations of long and complex texts and therefore leads to better performances in several text classification tasks.


MHS-RNN however suffers from some drawbacks. Following \cite{chung2016hierarchical}, the detection of a dynamic boundary or phrase in MHS-RNN is essentially done by treating the boundary detector as an extra gate unit in LSTM \cite{schmidhuber1991neural}. When a new dynamic boundary has been detected, the information will be passed to the phrase-layer through the gate and further update the phrase hidden state. Subsequently, the updated phrase hidden state needs to be passed back to the word-layer to start the processing of the next words. This procedure is necessary for general multi-scale RNNs, but is too complicated especially when interacting with static boundaries. Moreover, this constant exchange between the word- and phrase-layers makes it difficult to impose attention mechanisms to the word- and phrase-layers and train them separately. Instead, MHS-RNN combines the units of the word- and phrase-layers to form blocks and then add attention mechanisms to the blocks. Although MHS-RNN with added attention mechanism demonstrated improved performances, the full potential of attention mechanisms has not been fully realized. 


In order to overcome the drawbacks and further improve upon MHS-RNN, in this paper, we propose to treat the dynamic boundary detector as latent indicator at the word-layer. Specifically, each word is equipped with an indicator, which is assumed to be a Bernoulli random variable whose parameter only depends on the hidden state of the associated word. When an indicator takes on the value 1, it indicates the end of a phrase. Note that when the word-layer is processed from the beginning to the end, all of the indicators will be calculated, and the dynamic boundaries or phrases can be determined. This helps achieve certain separation between the processing of the word- and phrase-layers. The separation further enables us to impose separate attention mechanisms to the word- and phrase-layers. Together with the latent indicators, the attention mechanism imposed on the word-layer passes information from the word-layer to the phrase-layer. The details of the proposed architecture will be presented in Section II.D.

The latent indicators themselves can be considered a new layer associated with the word-layer, which we refer to as the indicator layer. Because the values of the indicators are unknown, the indicator layer is considered a layer with missing values. When training the proposed model, we apply the Expectation-Maximization (EM) algorithm \cite{mclachlan2007algorithm}
to handle the indicator layer. The indicator layer equipped with the EM algorithm not only simplifies the architecture of MHS-RNN, but also much improves its performance in text classification tasks. We refer to the proposed new architecture as the EM-HRNN model. 


In training, EM-HRNN faces one challenge. 
When calculating the $Q$ function in the EM algorithm, the computational complexity increases exponentially as the length of the text increases. In order to reduce the computational complexity, we propose two different bootstrap strategies. The first strategy is to divide a text into consecutive non-overlapping fragments, and then EM-HRNN is trained on those fragments one by one with the parameters of the other fragments fixed. We refer to this strategy as non-overlapping block bootstrap \cite{radovanov2014comparison}.
%
The second strategy uses the local block bootstrap method \cite{paparoditis2002local}. Experimental studies show that when using either of the two bootstrap strategies, our proposed EM-HRNN model outperforms most other RNN-based models. The performance of EM-HRNN with local block bootstrap is fairly close to some transformer-based models \cite{sun2019fine}.

The major contributions of this paper are summarized as follows. 
\begin{itemize}
    \item We propose to use latent indicators (i.e., the indicator layer) instead of dynamic boundary detectors to identify dynamic segments(e.g, phrases) in the usual mixed hierarchical structure of a text and further develop an EM algorithm to handle the indicator layer during training. 
    \item Taking an advantage of the latent indicator layer, we also impose attention mechanism to the dynamic layer (e.g, phrase layer), and therefore integrate a text's mixed hierarchical structure and attention mechanisms into a unified model called the EM-HRNN model. The EM-HRNN model demonstrates promising performances in simulation studies and real data applications. 
    \item To mitigate the computational complexity encountered when training the EM-HRNN model on long texts, we propose two bootstrap strategies, which are the non-overlapping block bootstrap method and the local block bootstrap method, respectively. Simulation studies and real data applications show that both strategies are able to train the EM-HMM model in an efficient and effective fashion.
\end{itemize}

\section{Model}
In this section, we first briefly review the basic structure of Long Short-Term Memory (LSTM). We further review the model of MHS-RNN with attention mechanism, discuss its major limitations, and present our ideas to improve upon the model. Then we apply the ideas and propose the Recurrent Neural Networks with Mixed Hierarchical Structures and EM Algorithm (in short, EM-HRNN). At last, we introduce efficient computational methods for training the proposed EM-HRNN model, which include an EM algorithm and two bootstrap algorithms.

\subsection{LSTM-base Sequence Encoder}
LSTM \cite{hochreiter1997long} was originally developed to address the issues of gradient vanishing and explosion in training vanilla RNNs for long sequences. Different from vanilla RNNs, LSTM uses gating mechanisms to track the states of sequences.
When updating the LSTM cell at time step $t$ of a sequence, the following calculations will be performed.

\begin{scriptsize}
\begin{equation}
\begin{split}
&i_{t} = \sigma(W_{i}h_{t-1} + U_{i}x_{t}+b_{i} ), \\
&f_{t} = \sigma(W_{f}h_{t-1} + U_{f}x_{t}+b_{f} ), \\
&\Tilde{c}_{t} = tanh(W_{c}h_{t-1} + U_{c}x_{t}+b_{c} ), \\
&o_{t} = \sigma(W_{o}h_{t-1} + U_{o}x_{t}+b_{o} ), \\
&c_t = i_t \odot \Tilde{c}_t + f_t \odot c_{t-1}, \\
&h_t = o_t \odot tanh(c_t). 
\end{split}
\end{equation}
\end{scriptsize}
Here $\sigma$ is the element-wise sigmoid function and $\odot$ is the element wise product; $x_t$ is the input vector at time $t$, and $h_t$ is the hidden-state vector at time $t$; $U_i$, $U_f$, $U_c$, and $U_o$ are the weight matrices of different gates for input $x_t$;  $W_i$, $W_f$, $W_c$, and $W_o$ are the weight matrices for hidden state $h_t$ at different gates. $b_i$, $b_f$, $b_c$, and $b_o$ denote the bias vectors. $f$, $i$, and $o$ correspond to the forget, input, and output gates of a LSTM cell.

\subsection{MHS-RNN with Attention}

Fig. 1 is a modified schematic diagram of the {\em MHS-RNN with attention} model. It contains three layers: a word-layer equipped with both static and dynamic boundary detectors, a phrase-layer generated from the word-layer by its dynamic boundary detector, and a sentence-layer generated from the phrase-layer by the static boundary detectors in the word-layer.

\begin{figure}[htbp]
  \centerline{\includegraphics[width=0.4\textwidth]{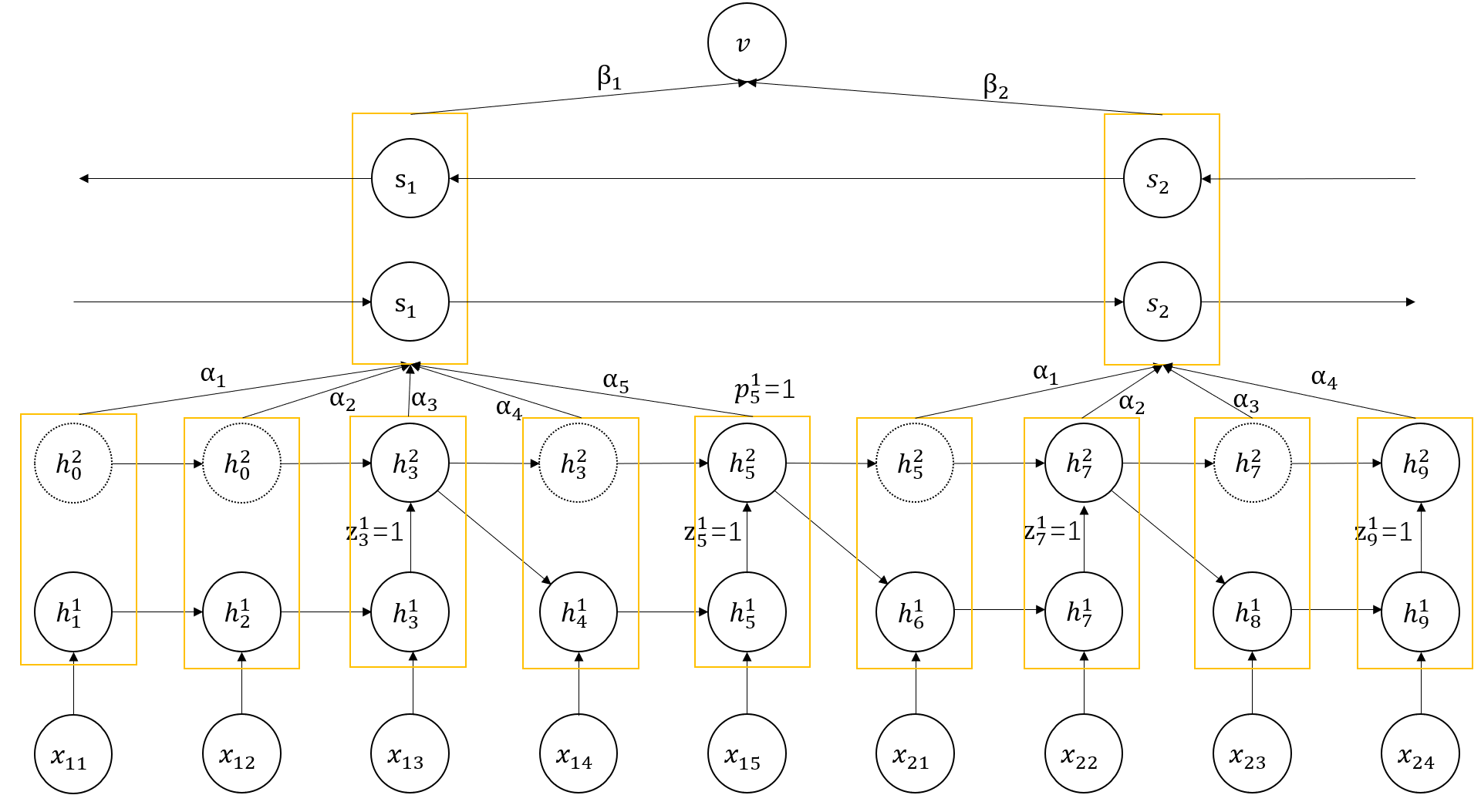}}
  \label{fig:xfig1}
  \caption{The MHS-RNN with attention architecture: $x_{ij}$ is the vector representation of $j$th input word in $i$th sentence, $h_i^j$ is the hidden state of time step $i$ layer $j$. The dotted cell indicates that there is no update here. We only marked the case where the detector is activated($ z$ or $ p = 1$ ).}
\end{figure}

There are two levels of attention mechanism structures in the model, which are represented by two layers of rectangles in Fig. 1. The first layer of rectangles contains the states of the word-layer and the phrase-layer, and is referred to as word-phrase attention. The word-phrase attention mechanism is used to extract the information from both the word- and phrase- layers, and aggregate them to form a sentence vector and pass it on to the sentence-layer. The second layer of rectangles appears in the sentence-layer in Fig. 1 and is referred to as sentence attention. The sentence attention is to reward sentences that provide important information of a document.

The model uses two types of boundary detectors in Fig. 1, where dynamic boundary detectors are denoted by $z^1_t$ and static boundary detectors are denoted by $p^1_t$. 
%
%
%
The static boundary detector $p^1_t$ is activated when punctuation marks are detected. In Fig. 1, $x_{15}$ and $x_{24}$ are the end of a sentence and the end of the document, respectively. Thus the static boundary detector is activated at $x_{15}$ and $x_{24}$ (i.e. $p^1_5=1$ and $p^1_9=1$), and the model will start to input the states of the word-layer and phrase-layer into the word-phrase attention mechanism and extract the sentence vector.

The dynamic boundary detector $z^1_t$ is used to detect dynamic boundaries that indicate the ends of phrases. When the end of a phrase segment is detected, the dynamic boundary detector will be turned on (e.g., $z^1_3=1$, $z^1_5=1$, $z^1_7=1$, and $z^1_9=1$ in Fig. 1), and the model will feed the state of the detected segment from the word-layer into the phrase-layer.
Whether the dynamic boundary detector is turned on or not is determined as $z_t^1 = 1$ if $\widetilde{z}^1_t > 0.5$ and $z_t^1 = 0$ otherwise. Here $\widetilde{z}^1_t$ is calculated by:

\begin{scriptsize}
\begin{equation}
\begin{split}
&   \widetilde{z}^1_t= hardsigm((1-z^1_{t-1})W_d h_{t-1}^1 + U_d x_t + z^1_{t-1} W_d h_{t-1}^2 + b_d)
\end{split}
\end{equation}
\end{scriptsize}
Where $hard sigm(x)=max (0, min (1, \frac{ax+1}{2}))$ with $a$ being the hyper-parameter {\em slope}, $W_d$ is the weight matrix for hidden state $h_t$, $U_d$ is the weight matrix for input $x_t$, and $b_d$ is the bias vector.

Notice that at time step $3$ in Fig. 1, the dynamic boundary detector is turned on (i.e., $z_3^1=1$), and the state $h_3^1$ is passed to the phrase-layer. Next, the model needs to reinitialize the state of $h_4^1$ with $h_3^2$ when learning $h_4^1$. In other words, the state of the phrase-layer ($h_3^2$) is passed back to the word-layer. 
This special operation happens whenever a dynamic boundary is detected during training.

\subsection{ Replacing Dynamic Boundary Detector with Latent Indicators}

From Equation (3), we can see that when calculating the dynamic boundary detector at time step $t$, it is necessary to consider $z^1_{t-1}$ at the previous time step $t-1$ as well as the states $h^1_{t-1}$ and $h^2_{t-1}$ of the word- and phrase-layers, respectively. This requires the model to simultaneously update the word- and phrase-layers at all time steps, which is not only difficult to execute during training, but also makes it difficult to incorporate attention mechanisms to the word- and phrase-layers separately. 
The reason is that the MHS-RNN model needs to refer to $h_{t-1}^2$ when updating $h_t^1$, whereas the attention mechanism needs to obtain all the states of the word-layer in order to calculate $h_{t-1}^2$. This clearly leads to a conflict. 
Again from Equation (3), it is clear that the dynamic boundary detector resembles the other gates in the LSTM cell, and is indeed more complicated because it involves both of the word- and phrase-layers. 


We believe that the dynamic boundary detector equipped with gate-like updating mechanism is over-complicated for detecting phrases. In particular, the feedback from the phrase-layer to the word-layer is unnecessary. Although phrases are not pre-annotated, they can be considered a latent structure embedded in the word-layer. In this paper, we propose to assign an indicator to each token of a text, which indicates whether the token is the end of a phrase. All the indicators together form a latent layer of the word-layer, and only depend on the hidden states of the word-layer. We refer to such a layer as the latent indicator layer. 
Unlike the dynamic boundary detector in the MHS-RNN model, the latent indicator layer does not depend on the phrase-layer, therefore any feedback from the phrase-layer to the word-layer. This greatly simplifies the model structure and computational complexity, and furthermore, it allows the incorporation of attention mechanisms to the word- and phrase-layers separately. 



Using the latent indicator layer, we integrate three levels of attention mechanisms into the hierarchical structure (i.e. the word-, phrase-, and sentence-layers) of a text and call the resulting architecture as the Recurrent Neural Network with Mixed Hierarchical Structure and EM Algorithm (in short, EM-HRNN). Here the EM algorithm \cite{mclachlan2007algorithm} refers to the computational method needed to handle the latent indicator layers during training. The latent indicators are not directly observable and can be considered missing values. During training, the EM algorithm can be used to impute the values of the indicators. We will present the EM-HRNN model, the EM algorithm, and 
additional computational methods in the next subsection.



\subsection{Model Architecture of EM-HRNN}

The architecture of the EM-HNN model is shown in Fig.2 and Fig.3. It consists of a number of layers: a word layer, an indicator layer, a word-level attention layer, a phrase layer, a phrase-level attention layer, a sentence layer, a sentence-level attention layer, and at last an output layer. We provide more details of these layers below.

\begin{figure}[htbp]
  \centerline{\includegraphics[width=0.4\textwidth]{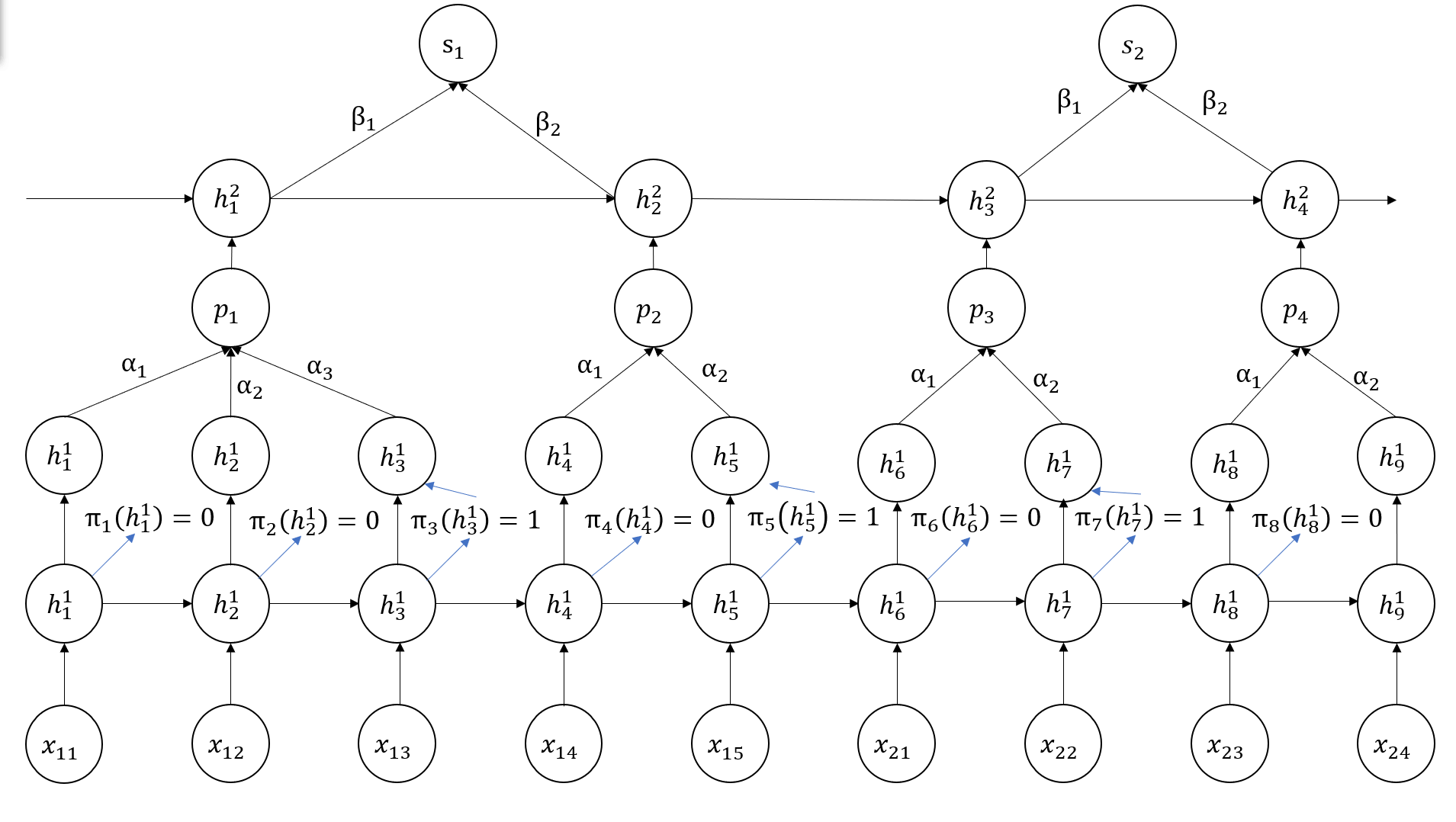}}
  \label{fig:xfig2}
  \caption{The first part of the EM-HRNN model. $x_{ij}$ is the word embedding obtained by pre-trained word2vec model. $z_i(h_i^1)$ is the value of the latent variable obtained by the indicator layer. $\alpha_i$'s and $\beta_i$'s are the attention weights with respect to the word-layer and the phrase-layer, respectively.}
\end{figure}

\begin{figure}[htbp]
  \centerline{\includegraphics[width=0.4\textwidth]{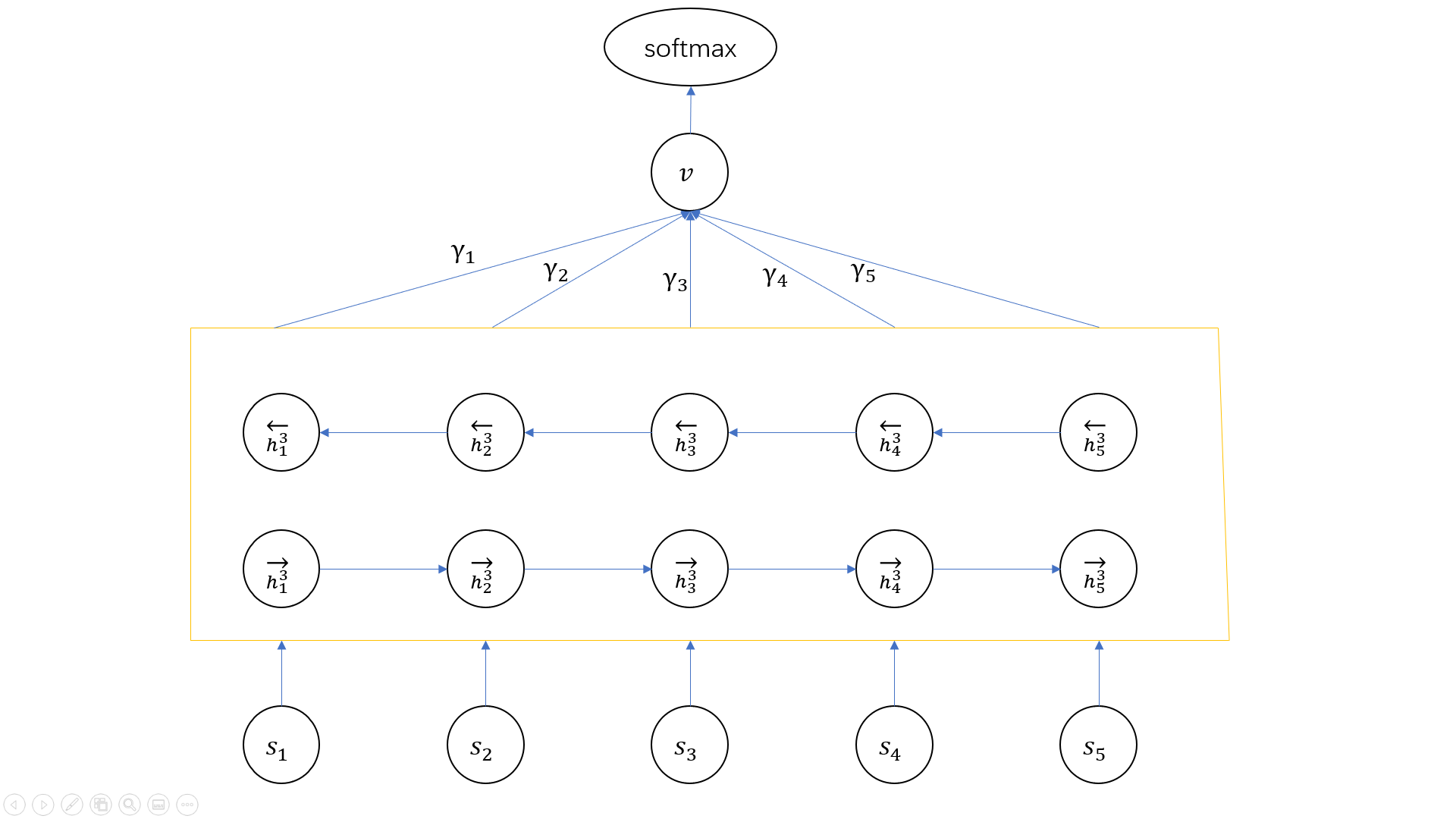}}
  \label{fig:xfig3}
  \caption{The second part of the EM-HRNN model. $s_i$ are the sentence vector obtained in the first part. $\gamma_i$'s are the attention weights with respect to sentence-layer. $v$ is the document vector calculated by the weighted sum of these units.}
\end{figure}

\textbf{Word layer} Assume a document has $L$ sentences denoted as $s_1$, $s_2$,$\dots$, $s_L$, respectively, and $w_{ij}$ representing the $j$th word in the $i$th sentence for $i$ = $1$, $2$, $\dots$, $L$, and $j=1, 2, \ldots, T_i$. First, we use the pre-trained word2vec \cite{mikolov2013efficient} model from GLOVE\footnote{https://nlp.stanford.edu/projects/glove/}to embed the word $w_{ij}$ and denote the result as $x_{ij}$, that is, $x_{ij} = Word2vec(w_{ij})$. Then we apply LSTM to process the word embeddings $w_{ij}$'s to obtain their annotations (i.e., hidden state $h_k^1$) as $h_k^1 = LSTM(x_{ij})$.

The separation between sentences will be directly processed in the word-layer, and the resulting information in this layer will be passed to phrase-layer and sentence-layer. This processing operation plays the same role as the static boundary detectors in MHS-RNN (i.e., $p_5=1$ in Fig. 2).

\textbf{Indicator layer}  As discussed in the previous subsection, we add a latent indicator layer on top of the world layer. Denote the indicator for the word at time step $t$ as $z_t$. Further, we assume that $z_t$ follows the Bernoulli distribution with intensity parameter $\pi_t$. When $z_t$ is turned on, that is, $z_t=1$, the corresponding word is considered to be the end of a phrase segment. $\pi_t$ is assumed to depend on the state of the word as $\pi_t = \sigma(W_{\pi} h_t^1 + b_{\pi})$. Therefore, we have $z_t = 1$ with probability $\pi_t$ and $z_t = 0$ with probability $1-\pi_t$.

Note that the indicators are not directly observable, that is, the exact value of $z_t$ is not available. In order to better present the remaining layers of the EM-HRNN model, we pretend that the values of $z_t$'s are known as in Fig. 1. During training, the values of $z_t$'s will be imputed by the EM algorithm as will be discussed later on. When $z_t=1$ at time step $t$, a phrase segment is detected, and the whole segment will be then fed to the word-attention layer. 


\textbf{Word-attention layer}  After the indicator layer divides all the words into segments. We add attention weights to the words in each segment to calculate the phrase vector and pass it on to the phrase-layer. We refer to this process as word-attention. 
For a segment, specifically,

\begin{scriptsize}
\begin{equation}
\begin{split}
&u_t^q = tanh(W_{q}h^1_{t} + b_{q} ), \\
&\alpha_{t} = \frac{exp((u_t^q)^Tu_q)}{\sum_t exp((u_t^q)^Tu_q)}, \\
&q_i = \sum_t \alpha_t h_t^1. 
\end{split}
\end{equation}
\end{scriptsize}
Here $W_q$ and $b_q$ are the weight matrix and bias vector for a Single-Layer Perceptron (SLP) and $u_q$ is a context vector. $q_i$ is the phrase vector that summarizes all the information of words in a phrase. We feed $h_t^1$ into the SLP and hence obtain a normalized importance weight $\alpha_t$ through a softmax function. After that, we calculate the phrase vector $p_i$ as a weighted sum of the concatenated vector $h_t^1$ based on the weights. Notice that the context vector $u_p$ is randomly initialized and jointly learned during the training process.

\textbf{Phrase layer}  After we obtain the phrase vectors (i.e., $p_i$'s), we apply LSTM to encode the phrase vectors as $h_t^2 = LSTM(q_t)$, and the outputs are then fed to the phrase-attention layer.

\textbf{Phrase-attention layer}  Next, we add attention weights to the encoded phrases and name this operation phrase-attention. The phrase-attention mechanism is used to extract the information from the phrase layer, and then aggregate them to obtain sentence vectors and further pass them on to the sentence-layer. Specifically,

\begin{scriptsize}
\begin{equation}
\begin{split}
&u_t^s = tanh(W_{s}h^2_{t} + b_{s} ), \\
&\beta_{t} = \frac{exp((u_t^s)^Tu_s)}{\sum_t exp((u_t^s)^Tu_s)}, \\
&s_i = \sum_t \beta_t h_t^2. 
\end{split}
\end{equation}
\end{scriptsize}
Here $s_i$ is the sentence vector that summarizes all the information of phrases in a sentence and $u_s$ is the context vector. 

\textbf{}{Sentence layer}  After we obtain the sentence vectors (i.e., $s_i$'s), we implement a bidirectional LSTM to encode the sentence vectors as $\overleftarrow{h_t^3} = \overleftarrow{LSTM}(s_t),\  \overrightarrow{h_t^3} = \overrightarrow{LSTM}(s_t)$.

We concatenate $\overleftarrow{h_t^3}$ and $\overrightarrow{h_t^3}$ to get an annotation 
$h_t^3 = [\overleftarrow{h_t^3},\overrightarrow{h_t^3}]$ of sentence $t$.

\textbf{Sentence attention layer}  At last, we add attention weights to the sentence annotation as shown in Fig. 3, and name this operation sentence attention. The sentence attention is to reward sentences that provide important information for a document. Specifically, 

\begin{scriptsize}
\begin{equation}
\begin{split}
&u_t^d = tanh(W_{d}h^3_{t} + b_{d} ), \\
&\gamma_{t} = \frac{exp((u_t^d)^Tu_d)}{\sum_i exp((u_t^d)^Tu_d)}, \\
&v = \sum_t \gamma_t h_t^3.
\end{split}
\end{equation}
\end{scriptsize}
Here $v$ is the document vector that summarizes all the information of sentences in a document and $u_d$ is the context vector. 
\textbf{Document Classification}  In the paper, we focus on the task of document classification. The document vector $v$ is a high-level representation of the document and can be used as features for document classification as $p = softmax(W_c v + b_c)$.

We use the negative log likelihood of the correct labels as training loss: $L = - \sum_d log \  p_{d_j}$, where $j$ is the label of document $d$.

\subsection{EM Algorithm}
The loss function $L$ above is in fact the complete likelihood function, under the assumption that the values of the latent indicators are known. In practice, as we mentioned in subsection $C$, they are not observable and thus missing. The Expectation-Maximization (EM) algorithm can be used to impute the latent indicators.

Consider a general statistical model $p(W, Z; \theta)$, in which $W$ represents the observed data, $Z$ the missing data, and $\theta$ the model parameters. Therefore, the complete likelihood function is $L(\theta; W, Z)$. The maximum likelihood estimate of $\theta$ denoted as $\hat{\theta}$ is defined as the maximizer of the marginal likelihood function
As $L(\theta;W) = P(W |\theta) = \int P(W,Z | \theta) dZ$ instead.


The EM algorithm calculates $\hat{\theta}$ by iteratively applying an Expectation step (E-step) and a Maximization step (M-step) as follows.
The E-step 
Calculates the expected log complete likelihood function under the current parameter estimate, 
${\theta^{(t)}}$: $Q(\theta | \theta^{(t)}) = E_{Z|W,\theta^{(t)}} [logL(\theta;W;Z)]$ and the
M-step Updates the parameter estimate by solving $\theta^{(t+1)} = arg\ max_\theta Q(\theta | \theta^{(t)})$

We apply the EM algorithm to the proposed EM-HRNN model as follows. Suppose the document under consideration is of length $n$. The indicators $z_1, z_2, \ldots, z_n$ are not observed and they form the missing data $Z=(z_1, z_2, \ldots, z_n)$. Recall the $\pi_1, \pi_2, \ldots, \pi_n$ are the intensity parameters of the indicators. Further, we use $\theta$ to represent the other parameters in the model. 

Assume that $\theta^{(i)}$ and  $\pi^{(i)}$ are the current estimates of the parameters. Then the $Q$ function for the EM algorithm can be defined as

\begin{scriptsize}
\begin{align*} \label{2}
& Q(\theta, \pi | \theta^{(i)} , \pi^{(i)}) \\
& = \sum_Z p(Z|y,w,\theta^{(i)},\pi^{(i)})log\ p(y,Z|w,\theta,\pi), \\
& = \sum_Z \frac{log\ p(y,Z|w,\theta,\pi) 
 p(y,Z|w,\theta^{(i)},\pi^{(i)})}{p(y|Z,w,\theta^{(i)},\pi^{(i)})} . \\
\end{align*}
\end{scriptsize}
Here $w$ is the vector of the input tokens, $y$ is the vector of the given labels for documents. We further simplify the $Q$ function and obtain a form that can be computed. Due to limited space, more details are omitted. Once $Q$ is available, we subsequently maximize $Q$ to update the parameter estimates.






Note that each time we update the Q function, we have to exhaust all the possible cases of the $n$ indicators, and the computational complexity is $2^n$. 
%
%
Therefore, as the sequence length increases, the calculation time increases exponentially. Next, we propose to use bootstrap methods to mitigate the computational cost.

\subsection{Bootstrap Strategies}



Similar to general time series data, correlation exists between consecutive tokens or words of a document. To preserve this correlation structure, we propose to use two block bootstrap strategies to train the EM-HRNN model. The two strategies are non-overlapping block bootstrap \cite{radovanov2014comparison} and local block bootstrap \cite{radovanov2014comparison}, respectively. 

Note that, when applying bootstrap, we choose the classification EM approach \cite{celeux1992classification} to impute the value of latent indicators, as $z_t = 1$ if $\pi_t > 0.5$ and $z_t = 1$ otherwise.

\textbf{Non-overlapping block bootstrap}  Non-overlapping block bootstrap divides the sequence data into several non-overlapping blocks. We then train the model on the blocks sequentially instead of directly train the model on the entire sequence. Note that when training a certain block, we will fix the parameters of other blocks.
%
The computational complexity to calculate Q function reduce from $2^n$ to $2^l \times \lceil \frac{n}{l} \rceil$ where $l$ is the length of the block.

\textbf{Local block bootstrap}  If the underlying stochastic structure is slowly changing with time, a local block-resampling procedure can be employed. Local block bootstrap selects several neighborhoods to form blocks and then train the model on those blocks. In this paper, We selected $10$ neighborhoods of length $5$.
The details of the local block bootstrap are shown in Algorithm 1.

\begin{algorithm}[htbp]
	\caption{Local block bootstrap} 
	\begin{algorithmic}[1]

		\For {$iteration\  i = 1,2,\ldots$,K},
		    	\For {$iteration\  i = 1,2,\ldots$,M},
	\State Randomly pick 10 tokens from the document (i.e., $\{ x_{i_1}, x_{i_2}, \dots ,x_{i_{10}}   \} \in \{ x_1, x_2, x_3, \dots , x_n  \}$). 
	\State  Create a neighbourhood of length $5$ for each select token to form a block. In this case, the $k$th block would be $B_k =\{ x_{i_k-2}, x_{i_k-1}, x_{i_k}, x_{i_k+1}, x_{i_k+2}  \}$.
			\State Update the parameter with respect to RNN network, fix the parameter with respect to latent indicator $Z$.
					\begin{scriptsize}
			\begin{align*}
         \theta^{(i+1)} = argmax_{\theta}\ Q(\theta, \pi=\pi^{(i)} | \theta^ {(i)},\pi=\pi^{(i)}).
         \end{align*}
         \end{scriptsize}
			\For {blocks $B_1,B_2,\ldots,B_{10}$}
				\State Update the parameter with respect to Block $B_k$, fixed other parameter.
				\begin{scriptsize}
				\begin{align*}
                  & \pi_{B_k}^{(i+1)} = \\ & argmax_{\pi_{B_k}} Q(\theta=\theta^{(i)}, \pi_{-B_k} 
                  =\pi_{-B_k}^{(i)}, 
                  \pi_{B_k}| \theta^ {(i)}, \pi^{(i)}).
\end{align*}	
\end{scriptsize}
			    \EndFor
		    \EndFor
		\EndFor
	\end{algorithmic} 
\end{algorithm}

In Algorithm 1, $\pi_{B_i}$ is the probability parameters $\pi$ with respect to block $B_i$ (i.e., $\pi_{B_i} = \{ \pi_{i_k-2}, \pi_{i_k-1}, \pi_{i_k}, \pi_{i_k+1}, \pi_{i_k+2}  \}$.  $\pi_{-B_i}$ denotes all the probability parameters exclude $\pi_{B_i}$.
For computing the Q function, the complexity reduces from $2^n$ to $2^5 \times 10 \times M$, where $M$ is the number of bootstrap samples.

\section{Experiment}
\subsection{Simulation Experiment}
The purpose of this experiment is to compare the capabilities of different hierarchical models for discovering hierarchical structures in simulated data.

We generate 10000 training documents and 1000 test documents. Each simulated document consists of two sentences, each of which consists of five tokens. The tokens here are represented by randomly generated 50-dimensional vectors, where we set the last token in each sentence to be the same vector. We then randomly generate the latent indicator $z$ for each token with the label of the last token in each sentence set to 1.

Since we know the ground truth of the segments of phrases is known, the documents can be fed into a three-layer attention LSTM with known parameters to generate document labels, where the three layers correspond to words, phrases and sentences, respectively, and the labels range from 1 to 5.

We compare the proposed EM-HRNN with two existing models MH-RNN \cite{chung2016hierarchical} and MHS-RNN \cite{luo2021recurrent}. In the experiment, we implement EM-HRNN with non-overlapping block bootstrap, local block bootstrap, and without bootstrap. 

Since the document length is 10, it is feasible to iterate over all possible segments. Thus we implement EM-HRNN without bootstrap which leads to the exact maximum likelihood estimates of the parameters. When implementing the non-overlapping block bootstrap strategy, We vary the block length from 1 to 5 as the sentence length is 5.

In the experiment, our focus is on how successful the models recover the latent indicator $z$'s. Note that for each document, there are 10 indicators. Thus overall we have 100000 indicators in the training dataset and 10000 indicators in the test dataset. The performance measure in this experiment is the percentage of indicators correctly recovered by a method.

\begin{table}[htpb]
\begin{center}
\scalebox{0.65}{
\begin{tabular}{|l|l|l|l|}
\hline 
\textbf{Method} & \textbf{Traing performance} & \textbf{Test performance} \\
\hline
HMRNN &  95.2\% &90.1\%  \\
MHS-RNN &  97.7\% &91.9\%  \\

EM-HRNN with window size 1 &  97.2\% & 89.68\%   \\
EM-HRNN with window size 2 &  97.2\% & 90\%   \\
EM-HRNN with window size 5 &  97.6\% & 91\%   \\
EM-HRNN with local bootstrap & 98\% &  92.7\%  \\
EM-HRNN without bootstrap & \textbf{98\%} & \textbf{93\% }\\
\hline
\end{tabular}}
\end{center}
\caption{Results of simulation experiment. EM-HRNN with window size $k$ represents EM-HRNN with non-overlapping bootstrap and block length $k$.}
\end{table}

From Table 1, we can see that EM-HRNN without bootstrap achieves the best performance as we expected while EM-HRNN with local bootstrap follows by a small margin. EM-HRNN with window sizes 1, 2 and 5 underperform MHS-RNN. Notice that when we increase the block length, the performance also increases. If we continue to increase the block length $k$, 
EM-HRNN with non-overlapping bootstrap will eventually outperform MHS-RNN but this also greatly increases the amount of computation. Taking into account both the amount of computation and performance, EM-HRNN with local bootstrap would be the best choice in practice.


In the following subsections, we will compare EM-HRNN with other existing document classification models in real datasets.

\subsection{Real Data Analysis}

\textbf{Datasets}  We evaluate our proposed model on five different document classification datasets. There are three datasets of Yelp reviews, which are obtained respectively from 2013, 2014, and 2015 Yelp dataset challenges. The other two are Amazon review and Yahoo answer. Among them, the Yelp reviews and Amazon reviews are sentiment classification tasks. Their labels range from 1 to 5, respectively, indicating that reviewers are very dissatisfied to very satisfied. The Yahoo answer is a topic classification task. There are ten topic classes in the Yahoo answer dataset. Details of these datasets can be found in related references.

\textbf{Settings and Details}  In the experiments, we set the dimension of the pre-trained word embedding method to be 100 following \cite{luo2021recurrent}. We only retain words that appear in the word2vec model and replace the other words with the special token {\em 'UNK'}.

The hyper-parameters are tuned on validation datasets. During experiments, we set the dimensions of all the involved layers to be 50 (following \cite{yang2016hierarchical}). We require the three attention mechanisms to have the same dimensions as the layers in the neural networks. Furthermore, we apply random initialization to all the layers.

For training, we set a mini-batch size to be 64 and organize documents of similar lengths to be batches. We use stochastic gradient descent to train all the models with a momentum of 0.9. Because the original datasets do not include the validation set, we randomly select $10\%$ of the training samples as the validation sets. We pick the best learning rate on the validation sets. 

For non-overlapping block bootstrap strategy, we increase the block length from 1 to 5. Note that when the block length is greater than 5, the amount of calculation increases significantly and the strategy become impractical.

\textbf{Results and Analysis}
The results are displayed in Table 2. 

\begin{table}[htbp]
\begin{center}
\scalebox{0.48}{
\begin{tabular}{|l|l|l|l|l|l|}
\hline \textbf{Methods} & \textbf{Yelp'13} & \textbf{Yelp'14} & \textbf{Yelp'15} & \textbf{Yahoo Answer} & \textbf{Amazon} \\ \hline
Bag-of-means \cite{zhang2015character} & - & - & 52.5 &  60.5  & 44.1 \\
SVM+SSWE \cite{tang2015document}& 53.5 & 54.3 & 55.4 &  -  & - \\
LSTM \cite{zhang2015character} & - & - & 58.2 &  70.8  & 59.4 \\
CNN-word \cite{zhang2015character} & - & - & 60.5 &  71.2  & 57.6 \\
Conv-GRNN \cite{tang2015document} & 63.7 & 65.5 & 66 &  -  & - \\
LSTM-GRNN \cite{tang2015document} & 65.1 & 67.1 & 67.6 &  -  & - \\
CMA \cite{ma2017cascading} & 66.4 & 67.6 & - & - & - \\
BiLSTM+linear-basis-cust \cite{kim2019categorical} & - & 67.1 & - & - & - \\
HN-AVE \cite{yang2016hierarchical} & 65.6 & 67.3 & 67.8 &  71.8  & 59.7 \\
HN-ATT \cite{yang2016hierarchical} & 66 & 68.9 & 69.4 &  73.8  & 60.7 \\
HM-RNN \cite{chung2016hierarchical} & 64 & 64.5 & 64.9 &  71  & 59 \\ 
MHS-RNN \cite{luo2021recurrent} & 65.2 & 67.5 & 67.7 &  72.3  & 59.7 \\
MHS-RNN with attention \cite{luo2021recurrent} & \textbf{66.8}& \textbf{69.3} & 69.9 &  74.1  & 61.2 \\ \hline
Bert-base \cite{sun2019fine} & - &  - & \textbf{71.4} &  \textbf{75.4}  & \textbf{61.9} \\ \hline
EM-HRNN with window size 1 & 66.8 & 69.2 & 70.5 &  74.4  & 61.4 \\
EM-HRNN with window size 2 & 66.9 & 68.3 & 70.8 &  74.7  & 61.6 \\
EM-HRNN with window size 5 & 67.1 & 69.5 & 71.2 &  75.1  & 61.9 \\ 
EM-HRNN with local bootstrap & \textbf{67.7}& \textbf{70.1} & \textbf{71.6} &  \textbf{75.7}  & \textbf{62.2} \\

\hline
\end{tabular}}
\end{center}
\caption{\label{font-table} Results in real datasets. The number here represents the prediction accuracy of the document label in the test set. Each dash lines in the table indicates that the corresponding dataset has not been reported by the reference paper. }

\end{table}

From Table 2, we compared our methods with both pre-trained classifier Bert-base and other non pre-trained classifiers. EM-HRNN with the non-overlapping block bootstrap with window size equal to 5 outperforms the existing non pre-trained best baseline classifiers by margins of 0.3, 0.2, 0.6, 0.6, and 0.7 in percentage points, respectively.

EM-HRNN with local bootstrap lead to even better results. On all datasets, {\em EM-HRNN with local bootstrap} outperforms the existing non pre-trained best baseline classifiers by margins of 0.9, 0.8, 1.7, 1.6, and 1 in percentage points, respectively.

The results demonstrate that by treating the phrase segmentation boundaries as latent indicators and incorporating the EM algorithm, EM-HRNN get a better performance in the document classification tasks. Bootstrap with local bootstrap strategy is an effective method to reduce computational complexity under the premise of less impact on performance.

Moreover, we compare to the EM-HRNN with a large pre-trained model called Bert-base \cite{sun2019fine}. In \cite{sun2019fine}, the authors fine tuned Bert-base on Yahoo answer dataset. Following this work, we fine tuned Bert-base on Yelp'15 and Amazon datasets and compare it to EM-HRNN. As shown in the Table 2, EM-HRNN with local bootstrap outperforms Bert-base by margins of 0.3, 0.2 and 0.3 in percentage points, respectively. The improvements of EM-HRNN over Bert-base is significant because Bert-base trained use more datasets for pretraining, and has ten times more parameters than EM-HRNN.

\subsection{Analysis of Latent Indicators}
Recall that latent indicator $Z$'s are not pre-annotated, instead, they are learned during the training of the models.
Not only can the latent indicator $Z$'s help produce better performances in document classification tasks, but they also provide linguistically meaningful segmentation of a text. In this subsection, we analyze the latent indicator $Z$'s in the experiments. Due to space limitations, here we choose to report the comparison between EM-RNN with local block bootstrap and MHS-RNN.

We first checked the lengths of the learned phrases. The length of a phrase is defined as the number of words between two active $Z$'s (i.e., $z = 1$) with the first active $z$ excluded and the second active $z$ included.

When comparing EM-HRNN with MHS-RNN, we can see that EM-HRNN prefers shorter segment phrases. Among all datasets, the average length of MHS-RNN phrase segmentation is 4.72, while that of EM-HRNN is 4.31. The minimum phrase length obtained by both of the methods is 1. The longest phrase obtained by MHS-RNN is 17, while that by EM-HRNN is 11.

We then randomly selected 200 documents from all 5 datasets. Among them, we selected two representative examples, given in Fig. 4 and Fig. 5.

\begin{figure}[htbp]
  \centerline{\includegraphics[width=0.45\textwidth,]{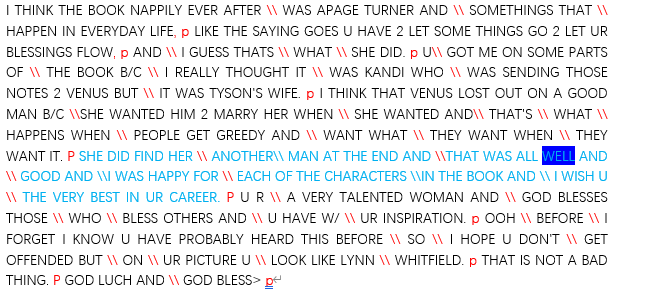}}
  \label{fig:xfig18}
  \caption{A sample from Amazon review dataset using MHS-RNN with attention. The double slash indicates that this place is a phrase clause obtained by the model. The colored part of the figure represents the highest proportion of the attention mechanism in the model.}
\end{figure}

\begin{figure}[htbp]
  \centerline{\includegraphics[width=0.45\textwidth,]{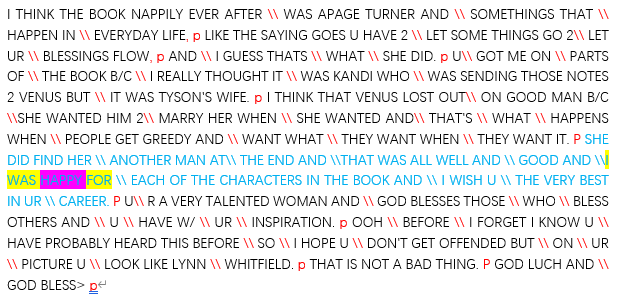}}
  \label{fig:xfig19}
  \caption{A sample from Amazon review dataset using EM-HRNN with local bootstrap. The double slash indicates that this place is a phrase clause obtained by the model. The colored part of the figure represents the highest proportion of the attention mechanism in the model.}
\end{figure}

One noteworthy place in the two figures is a clause, located between the first punctuation mark and the second punctuation mark: {\em LIKE THE SAYING GOES U HAVE 2 LET SOME THINGS GO 2 LET UR BLESSINGS FLOW}. In MHS-RNN, this clause is not segmented into phrases. But in EM-HRNN, this sub-sentence is segmented into {\em LIKE THE SAYING GOES U HAVE 2}, {\em LET SOME THINGS GO 2}, {\em LET UR} and {\em BLESSINGS FLOW}. Note that {\em 2} here is the abbreviation of {\em to}, and {\em UR} in this place is the abbreviation of {\em you are}. 
The phrase segmentations from EM-HRNN is more detailed and meaningful than MHS-RNN. Combined with the fact that EM-HRNN is more inclined to produce shorter phrases. We believe that the phrases obtained by EM-HRNN are closer to a semantic unit.

We then explore the quality of the attention mechanisms used in EM-HRNN and MHS-RNN. We compare the attention weights in both models on some words that express strong emotions.

\begin{table}[htbp]
\begin{center}
\scalebox{0.75}{
\begin{tabular}{|l|l|l|}
\hline 
\textbf{Words} & \textbf{Avg weights in MHS-RNN} & \textbf{Avg weights in EM}  \\
\hline
Good & 0.62 & 0.69   \\
Bad & 0.57 & 0.67   \\
Great & 0.51 & 0.56   \\
Sad & 0.49 & 0.54 \\
Excellent & 0.44 & 0.51  \\
\hline
\end{tabular}}
\end{center}
\caption{\label{table1} The average attention of some commonly used words that indicate strong emotions.}
\end{table}

It can be seen from the Table 3 that among all these words with strong emotions, the attention weights assigned by EM-HRNN are overall higher than those of MHS-RNN. We next made a detailed comparison among the 200 randomly selected samples mentioned above. We compared the most important words in the documents selected by the two models. Here the most important word in a document is obtained by following steps. For HM-RNN, we first find the sentence with most attention weights. Then we find the phrase with most attention weights in that sentence. At last we find the word with most attention weights in that phrase. For MHS-RNN, we first find the sentence with most attention weights. Then we directly find the words with most attention weights in that sentence.

Among all 200 documents, the two models selected the same important words in 176 documents and selected differently in the other 24 documents. Among them, we believe that the words selected by EM-HRNN in 14 documents can better express the emotional tendency of the documents. There are 7 documents in which the two models have selected different words, and we cannot determine which one is better. There are only 3 documents, in which we think the words selected by MHS-RNN are more representative.

In the two examples we showed in Figure 3 and 4 earlier, both models assign the largest attention weight to sentence {\em SHE DID FIND HER ANOTHER MAN AT THE END AND THAT WAS ALL WELL AND GOOD AND I WAS HAPPY FOR EACH OF THE CHARACTERS IN THE BOOK AND I WISH U THE VERY BEST IN UR CAREER}. But the word-phrase attention in MHS-RNN assigns the maximum attention to the word {\em well} in the phrase {\em that was all well and}. But the attention mechanism in EM-HRNN assigns the highest attention weight to the phrase {\em I was happy for} and assigns the highest attention weight to the word {\em happy} in the above phrase.
Consider the entire paragraph, we can see that the phrase 
{\em I was happy for} has a stronger sentiment than the phrase {\em that was all well and}, which is more helpful to judge the sentiment of the document. We believe that the hierarchical attention mechanism composed of words, phrases, and sentences achieves a better performance.

\section{Conclusion}
In the paper, we propose the EM-HRNN model to represent both explicit and implicit hierarchical information in a text, and further integrate different levels of attention mechanisms into the model. Using bootstrap strategies, the EM-HRNN model outperforms other RNN-based hierarchical models in document classification tasks and also demonstrates better performance than Bert-base. There are two directions we will pursue to further investigate the proposed model. First, we will study the performances of the EM-HRNN model in other NLP tasks. Second, a more thorough comparison study with other Transformer-based models is needed in order to understand the potential of the EM-HRNN model.

\section{Bibliographical References}\label{reference}

\bibliographystyle{lrec2022-bib}
\bibliography{lrec2022-example}
\end{document}